# Whale swarm algorithm for function optimization

Bing Zeng
School of Mechanical Science
and Engineering
Huazhong University of Science
and Technology
Wuhan, China
zengbing2016@126.com

Liang Gao*
School of Mechanical Science
and Engineering
Huazhong University of Science
and Technology
Wuhan, China
gaoliang@hust.edu.cn

Xinyu Li
School of Mechanical Science
and Engineering
Huazhong University of Science
and Technology
Wuhan, China
lixinyu@hust.edu.cn

*Abstract*—Increasing nature-inspired metaheuristic algorithms are applied to solving the real-world optimization problems, as they have some advantages over the classical methods of numerical optimization. This paper has proposed a new nature-inspired metaheuristic called Whale Swarm Algorithm for function optimization, which is inspired by the whales' behavior of communicating with each other via ultrasound for hunting. The proposed Whale Swarm Algorithm has been compared with several popular metaheuristic algorithms on comprehensive performance metrics. According to the experimental results, Whale Swarm Algorithm has a quite competitive performance when compared with other algorithms.

## I. INTRODUCTION

Nature-inspired algorithms are becoming powerful in solving numerical optimization problems, especially the NP-hard problems such as the travelling salesman problem [1], vehicle routing [2], classification problems [3], routing problem of wireless sensor networks(WSN) [4] and multiprocessor scheduling problem [5], etc. These real-world optimization problems often probably come with multiple global or local optima of a given mathematical model (i.e., objective function). And if a point-by-point classical method of numerical optimization is used for this task, the classical method has to try many times for locating different optimal solution in each time [6], which will take a lot of time and work. Therefore, using nature-inspired metaheuristic algorithms to solve these problems has become a hot research topic, as they are easy to implement and can converge to the global optima with high probability. In this paper, we have proposed a new nature-inspired metaheuristic called Whale Swarm Algorithm (WSA) for function optimization, based on the whales' behavior of communicating with each other via ultrasound for hunting. Here, a brief overview of the nature-inspired metaheuristic algorithms is presented.

Genetic Algorithm (GA) was initially proposed by Holland to solve the numerical optimization problem [7], which simulates Darwin's genetic choice and natural elimination biology evolution process and has opened the prelude of nature-inspired metaheuristic algorithms. It mainly utilizes selection, crossover and mutation operations on the individuals (chromosomes) to find the global optimum as far as possible. In GAs, the crossover operator that is utilized to create new individuals by combining parts of two individuals significantly affects the performance of a genetic system [8]. Until now, lots of researchers have proposed diverse crossover operators for different optimization problems. For instance, Syswerda has proposed order based crossover operator (OBX) for permutation encoding when dealing with schedule optimization problem [9]. A detailed review of crossover operators for permutation encoding can be seen from reference [10]. Mutation is another important operator in GAs, which provides a random diversity in the population [11], so as to prevent premature convergence of algorithm. Michalewicz has proposed random (uniform) mutation and non-uniform mutation [12] for numerical optimization problems. And polynomial mutation operator proposed by Deb is one of the most widely used mutation operator [13]. A comprehensive introduction to mutation operator can be seen from [14]. In a word, it is very important to choose or design appropriate select, crossover and mutation operators of GAs, when dealing with different optimization problems.

Storn and Price proposed Differential Evolution (DE) algorithm for minimizing possibly nonlinear and non-differentiable continuous space functions [15]. It also contains three key operations, namely mutation, crossover and selection, which are different from those of GAs. First of all, a donor vector, corresponding to each member vector of the population called target vector, is generated in the mutation phase of DE. Then, the crossover operation takes place between the target vector and the donor vector, wherein a trial vector is created by selecting components from the donor vector or the target vector with the crossover probability. The selection process determines whether the target or the trial vector survives in the next generation. If the trial vector is better, it replaces the target vector; otherwise remaining the target vector in the population. Since put forward, DE algorithm has gained increasing popularity from researchers and engineers in solving lots of real-world optimization problems [16, 17] and various schemes have been proposed for it [18]. The general convention used to name the different DE schemes is "DE/*x*/*y*/*z*", where DE represents "Differential Evolution", *x* stands for a string indicating the base vector need to be perturbed, for example, it can be set as "*best*" and "*rand*", *y* denotes the number of difference vectors used to perturb *x*, and *z* represents the type of crossover operation which can be binomial (*bin*) or exponential (*exp*) [18]. Some popular existing DE schemes are DE/*best*/1/*bin*, DE/*best*/1/*exp*, DE/*rand*/1/*bin*, DE/*best*/2/*exp*, E/*rand*/2/*exp*, etc.

Particle Swarm Optimization (PSO) is a swarm intelligence based algorithm proposed by Kennedy and Eberhart, which is inspired by social behavior of bird flocking [19]. PSO algorithm has been applied to solve lots of complex and difficult real-world optimization problems [20, 21], since it was put forward. In the traditional PSO algorithm, each particle moves to a new position based on the update of its velocity and position, where the velocity is concerned with its cognitive best position and social best position. Until now, there are lots of PSO variants are proposed for different optimization problems. For instance, Shi and Eberhart has introduced a linear decreasing inertia weight into PSO (PSO-LDIW) [22], which can balance the global search and local search, for function optimization. Zhan et al. have proposed Adaptive PSO (APSO) [23] for function optimization, which enables the automatic control of parameters to improve the search efficiency and convergence speed, and employs an elitist learning strategy to jump out of the likely local optima. Qu et al. have proposed Distance-based Locally Informed PSO (LIPS) that eliminates the need to specify any niching parameter and enhance the fine search ability of PSO for multimodal function optimization [6], etc.

In addition to the above, there are large amounts of other nature inspired algorithms such as Ant Colony Optimization (ACO) [24], Bees Swarm Optimization (BSO) [25] and Big Bang-Big Crunch (BB-BC) [26], etc. A comprehensive review of the nature inspired algorithms is beyond the scope of this paper. A detailed and complete reference on the motif can be seen from [27, 28].

The rest of this paper is organized as follows. Section 2 describes the proposed WSA in sufficient detail. The experiment setup is presented in Section 3. Section 4 presents the experimental results performed to evaluate the proposed algorithm. The last section is the conclusions and topics for further works.

## II. WHALE SWARM ALGORITHM

First of all, this section introduces the behavior of whales probably, especially the behavior of whales hunting. Then, the details of Whale Swarm Algorithm are presented.

### A. Behavior of whales

Whales with great intellectual and physical capacities are completely aquatic mammals, and there are about eighty whale species in the vast ocean. They are social animal and live in groups. Such as pregnant females will gather together with other female whales and calves so as to enhance defense capabilities. And sperm whales are often spotted in groups of some 15 to 20 individuals, as shown in Fig. 1. The whale sounds are beautiful songs in the oceans and their sound range is very wide. Until now, scientists have found 34 species of whale sounds, such as whistling, squeaking, groaning, longing, roaring, warbling, clicking, buzzing, churring, conversing, trumpeting, clopping and so on. These sounds made by whales can often be linked to important functions such as their migration, feeding and mating patterns. What's more, a large part of sounds made by whales are ultrasound which are beyond the scope of human hearing. And whales determine foods azimuth and keep in touch with each other from a great distance by the ultrasound.

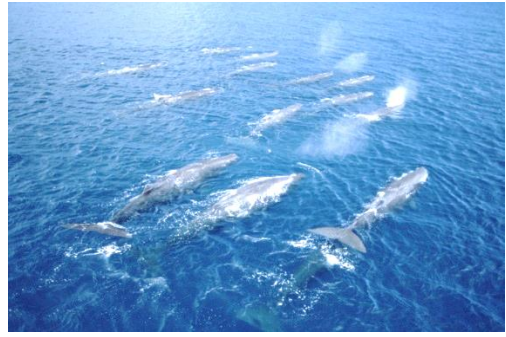

Fig. 1. The swarm of sperm whales.

When a whale has found food source, it will make sounds to notify other whales nearby of the quality and quantity of food. So each whale will receive lots of notifications from the neighbors, and then move to the proper place to find food based on these notifications. The behavior of whales communicating with each other by sound for hunting inspire us to develop a new metaheuristic algorithm for function optimization problems. In the rest of this section, we will discuss the implementation of Whale Swarm Algorithm in detail.

### B. Whale swarm algorithm

To develop whale swarm inspired algorithm for solving function optimization problem, we have idealized some hunting rules of whale. For simplicity in describing our new Whale Swarm Algorithm, the following four idealized rules are employed: 1) all the whales communicate with each other by ultrasound in the search area; 2) each whale has a certain degree of computing ability to calculate the distance to other whales; 3) the quality and quantity of food found by each whale are associated to its fitness; 4) the movement of a whale is guided by the nearest one among the whales that are better (judged by fitness) than it, such nearest whale is called the "better and nearest whale" in this paper.

#### 1) Iterative equation

As we know, both radio wave and light wave are electromagnetic waves, which can propagate without any medium. If propagating in water, they will attenuate quickly due to the large electrical conductivity of water. Whereas, sound wave is one kind of mechanical wave that needs a medium through which to travel, whether it is water, air, wood or metal. And ultrasound belongs to sound wave, whose transmission speed and distance largely depends on the medium. For instance, ultrasound travels about 1450 meters per second in water, which is faster than that (about 340 meters per second) in air. What's more, some ultrasound with pre-specified intensity can travel about 100 meters underwater, but can only transmit 2 meters in air. That is because the intensity of mechanical wave is continuously attenuated by the molecules of the medium, and the intensity of ultrasound traveling in air is attenuated far more quickly than that in water. The intensity $\rho$ of the ultrasound at any distance $d$ from the source can be formulated as follows [29].

$$\rho = \rho_0 \cdot e^{-\eta \cdot d} \quad (1)$$

where, $\rho_0$ is the intensity of ultrasound at the origin of source, $e$ denotes the natural constant. $\eta$ is the attenuation coefficient, which depends on the physico-chemical properties of the medium and on the characteristics of the ultrasound itself (such as the ultrasonic frequency) [29].

As we can see from Eq. 1, $\rho$ decreases exponentially with the increment of $d$ when $\eta$ is constant, which means that the distortion of message conveyed by the ultrasound transmitted by a whale will occur with a great probability, when the travel distance of the ultrasound gets quite far. So a whale will not sure whether its understanding of the message send out by another whale is correct, when that whale is quite far away from it. Thus, a whale would move negatively and randomly towards its better and nearest whale which is quite far away from it.

Based on the above, it can be seen that a whale would move positively and randomly towards its better and nearest whale which is close to it, and move negatively and randomly towards that whale which is quite far away from it, when hunting food. Thus, some whale swarms will form after a period of time. Each whale moves randomly towards its better and nearest whale, because random movement is an important feature of whales' behavior, like the behavior of many other animals such as ant, birds, etc., which is employed to find better food. These rules have inspired us to find a new position iterative equation, wishing the proposed algorithm to avoid falling into the local optima quickly and enhance the population diversity and the global exploration ability, as well as contribute to locating multiple global optima. Then, the random movement of a whale $\mathbf{X}$ guided by its better and nearest whale $\mathbf{Y}$ can be formulated as follows.

$$x_i^{t+1} = x_i^t + \text{rand}\left(0, \rho_0 \cdot e^{-\eta \cdot d_{\mathbf{X},\mathbf{Y}}}\right) * \left(y_i^t - x_i^t\right) \qquad (2)$$

where, $x_i^t$ and $x_i^{t+1}$ are the $i$-th elements of $\mathbf{X}$'s position at $t$ and $t+1$ iterations respectively, similarly, $y_i^t$ denotes the $i$-th element of $\mathbf{Y}$'s position at $t$ iteration. $d_{\mathbf{X},\mathbf{Y}}$ represents the Euclidean distance between $\mathbf{X}$ and $\mathbf{Y}$. And $\text{rand}(0, \rho_0 \cdot e^{-\eta \cdot d_{\mathbf{X},\mathbf{Y}}})$ means a random number between 0 and $\rho_0 \cdot e^{-\eta \cdot d_{\mathbf{X},\mathbf{Y}}}$. Based on a large number of experiments, $\rho_0$ can be set to 2 for almost all the cases.

As mentioned previous, the attenuation coefficient $\eta$ is dependent on the physico-chemical properties of the medium and on the characteristics of the ultrasound itself. Here, for function optimization problem, those factors that affect $\eta$ can be associated to the characteristics of the objective function, including the function dimension, range of variables and distribution of peaks. Therefore, it is important to set appropriate $\eta$ value for different objective function. For engineer's convenience in application of WSA, the initial approximate value of $\eta$ can be set as follows, based on a large number of experimental results. First of all, we should make $\rho_0 \cdot e^{-\eta \cdot (d_{\max}/20)} = 0.5$, i.e., $2 \cdot e^{-\eta \cdot (d_{\max}/20)} = 0.5$, since $\rho_0$ is always set to 2, wherein $d_{\max}$ denotes the maximum distance between any two whales in the search space that can be formulated as $d_{\max} = \sqrt{\sum_{i=1}^{n}(x_i^U - x_i^L)^2}$, $n$ is the dimension of the objective function, $x_i^L$ and $x_i^U$ represent the lower limit and upper limit of the $i$-th variable respectively. This equation means that if the distance between whale $\mathbf{X}$ and its better and nearest whale $\mathbf{Y}$ is $d_{\max}/20$, the part $\rho_0 \cdot e^{-\eta \cdot d_{\mathbf{X},\mathbf{Y}}}$ of Eq. 2 that affects the moving range of whale $\mathbf{X}$ should be set to 0.5. Next, we can get that $\eta = -20 \cdot \ln(0.25)/d_{\max}$. Then, it is easy to adjust $\eta$ to the optimal or near-optimal value based on this initial approximate value.

Eq. 2 shows that a whale will move towards its better and nearest whale positively and randomly, if the distance between them is small. Otherwise, it will move towards its better and nearest whale negatively and randomly, which can be illustrated with Fig. 2 when the dimension of the objective function is equal to 2. In Fig. 2, the red stars denote the global optima, the circles represent the whales and the rectangular regions signed with imaginary lines are the reachable regions of the whales in current iteration.

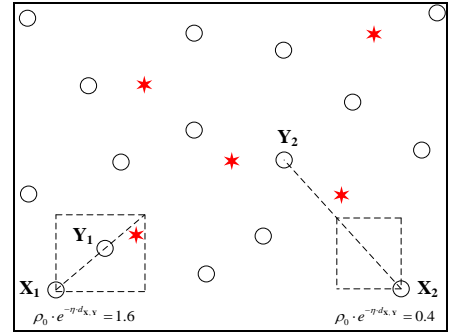

Fig. 2. Sketch map of a whale's movement guided by its better and nearest whale.

*2) General framework of WSA*

Based on the above rules, the general framework of the proposed WSA can be summarized as shown in Fig. 3, where $|\mathbf{\Omega}|$ in line 6 denotes the number of members in $\mathbf{\Omega}$, namely the swarm size, and $\mathbf{\Omega}_i$ in line 7 is the $i$-th whale in $\mathbf{\Omega}$. It can be seen from Fig. 3 that those steps before iterative computation are some initialization steps, including initializing configuration parameters, initializing individuals' positions and evaluating each individual, which are similar with most other metaheuristic algorithms. Here, all the whales are randomly assigned to the search area. Next come the core step of WSA: whales move (lines 5-13). Each whale needs to move for the better food via group cooperation. First of all, a whale should find its better and nearest whale (lines 7), as shown in Fig. 4, where $f(\mathbf{\Omega}_i)$ in line 6 is the fitness value of the whale $\mathbf{\Omega}_i$ and $dist(\mathbf{\Omega}_i, \mathbf{\Omega}_u)$ in line 7 denotes the distance between $\mathbf{\Omega}_i$ and $\mathbf{\Omega}_u$. If its better and nearest whale exists, then it will move under the guidance of the better and nearest whale (lines 9 in Fig. 3). As described above, the framework of WSA is fairly simple, which is convenient for applying WSA in solving the real-world optimization problems.

| The general framework of Whale Swarm Algorithm |
| --- |
| **Input:** An objective function, the whale swarm $\mathbf{\Omega}$. |
| **Output:** The global optima. |
| 1: **begin** |
| 2:     Initialize parameters; |
| 3:     Initialize whales' positions; |
| 4:     Evaluate all the whales (calculate their fitness); |
| 5:     **while** termination criterion is not satisfied **do** |
| 6:         **for** $i$=1 **to** $|\mathbf{\Omega}|$ **do** |
| 7:             Find the better and nearest whale $\mathbf{Y}$ of $\mathbf{\Omega}_i$; |
| 8:             **if** $\mathbf{Y}$ exists **then** |
| 9:                 $\mathbf{\Omega}_i$ moves under the guidance of $\mathbf{Y}$ according to Eq. 2; |
| 10:                Evaluate $\mathbf{\Omega}_i$; |
| 11:            **end if** |
| 12:         **end for** |

```
13: end while
14: return the global optima;
15: end
```

Fig. 3. The general framework of WSA.

```
The pseudo code of finding a whale's better and nearest whale
Input: The whale swarm Ω, a whale Ωᵤ.
Output: The better and nearest whale of Ωᵤ.
 1: begin
 2: Define an integer variable v initialized with 0;
 3: Define a float variable temp initialized with infinity;
 4: for i=1 to |Ω| do
 5:     if i≠u then
 6:         if f(Ωᵢ)<f(Ωᵤ) then
 7:             if dist(Ωᵢ, Ωᵤ)<temp then
 8:                 v=i;
 9:                 temp=dist(Ωᵢ, Ωᵤ);
10:             end if
11:         end if
12:     end if
13: end for
14: return Ωᵥ;
15: end
```

Fig. 4. The pseudo code of finding a whale's better and nearest whale.

## III. EXPERIMENT SETUP

### A. Experimental configuration

The proposed WSA and other algorithms compared are all implemented with C++ programming language by Microsoft visual studio 2015 and executed on the PC with 2.3 GHz Intel core i7 3610QM processor, 8 GB RAM and Microsoft Windows 10 operating system. In addition to the GA with elite selection, non-uniform arithmetic crossover and basic bit mutation strategies, DE/best/1/bin [30] and PSO with inertia weight [22], the following 4 popular multimodal optimization algorithms are also compared with WSA: The locally informed PSO (LIPS) [6], Speciation-based DE (SDE) [31], The original crowding DE (CDE) [32] and Speciation-based PSO (SPSO) [33].

In this paper, we utilize the evaluation number of objective function as the stopping criterion of these algorithms to test their performance.

### B. Test functions

To verify the performance of the proposed WSA, the comparative experiments were conducted on twelve benchmark test functions, which are taken from the studies of Deb [34], Michalewicz [12], Li [35] and Thomsen [32]. Test functions F1-F8 are multimodal (F1-F6 and F7-F10 are low and high dimensional multimodal functions respectively) that have multiple global or local optima. F11 and F12 are high dimensional unimodal functions with 100 dimension. Basic information of these test functions are summarized in Table 1. For functions F2-F6, the objective is to locate all the global optima, while for the rest the target is to escape the local optima (if they have) to hunt for the global optimum. And all test functions are minimization problems. It can be seen from Table 1, F7, F9, F11 and F12 all get the global optima at (0, 0, …, 0), and F10 gets the global optimum at (1, 1, …, 1), which are located near the middle of the feasible region. As we know, some algorithms are efficient in optimizing the functions whose optima are near the middle of the feasible region, especially near zero, but perform badly when the optima are not near the middle of the feasible region. For the sake of fairness, we have shifted F7-F12, and the shift data of them are randomly generated within specified range.

Table 1. Test functions.

| No. | Function name / Dimensions | Expression | Ranges | No. of global optima | Minimum value |
|---|---|---|---|---|---|
| 1 | Uneven Increasing Minima / 1D | $f(X) = -\exp\left(-2\log(2)\cdot\left(\frac{x_1-0.08}{0.854}\right)^2\right)\cdot\sin^6\left(5\pi\left(x_1^{3/4}-0.05\right)\right)$ | [0, 1] | 1 | -1 |
| 2 | Uneven Minima / 1D | $f(X) = -\sin^6\left(5\pi\left(x_1^{3/4}-0.05\right)\right)$ | [0, 1] | 5 | -1 |
| 3 | Himmelblau's function / 2D | $f(X) = \left(x_1^2+x_2-11\right)^2+\left(x_1+x_2^2-7\right)^2-200$ | [-6, 6]² | 4 | -200 |
| 4 | Six-hump camel back / 2D | $f(X) = 4\left(\left(4-2.1x_1^2+\frac{x_1^4}{3}\right)x_1^2+x_1x_2+\left(-4+4x_2^2\right)x_2^2\right)$ | [-1.9, 1.9] [-1.1, 1.1] | 2 | -4.126514 |
| 5 | inverted Shubert function / 2D | $f(X) = \prod_{i=1}^{2}\sum_{j=1}^{5}j\cos\left((j+1)x_i+j\right)$ | [-10, 10]² | 18 | -186.7309 |
| 6 | Branin RCOS / 2D | $f(X) = \left(x_2-\frac{5.1}{4\pi^2}x_1^2+\frac{5}{\pi}x_1-6\right)^2+10\left(1-\frac{1}{8\pi}\right)\cos(x_1)+10$ | [-5, 10] [0, 15] | 3 | 0.397887 |
| 7 | Rastrigin / 100D | $f(X) = \sum_{i=1}^{D}\left(x_i^2-10\cos(2\pi x_i)+10\right)$ | [-100, 100]¹⁰⁰ | 1 | 0 |
| 8 | Schwefel / 100D | $f(X) = D\cdot 418.9829-\sum_{i=1}^{D}x_i\sin\left(\sqrt{|x_i|}\right)$ | [-500, 500]¹⁰⁰ | 1 | 0 |
| 9 | Griewank / 100D | $f(X) = -\prod_{i=1}^{D}\cos\left(\frac{x_i}{\sqrt{i}}\right)+\sum_{i=1}^{D}\frac{x_i^2}{4000}+1$ | [-100, 100]¹⁰⁰ | 1 | 0 |
| 10 | Rosenbrock / 100D | $f(X) = \sum_{i=1}^{D-1}\left(100\left(x_{i+1}-x_i^2\right)^2+\left(x_i-1\right)^2\right)$ | [-15, 15]¹⁰⁰ | 1 | 0 |
| 11 | Sphere / 100D | $f(X) = \sum_{i=1}^{D}x_i^2$ | [-100, 100]¹⁰⁰ | 1 | 0 |
| 12 | Zakharov / 100D | $f(X) = \sum_{i=1}^{D}x_i^2+\left(\frac{1}{2}\sum_{i=1}^{D}ix_i\right)^2+\left(\frac{1}{2}\sum_{i=1}^{D}ix_i\right)^4$ | [-5, 10]¹⁰⁰ | 1 | 0 |

## C. Parameters setting

Although the global optima of these test functions can be obtained by the method of derivation, they should still be treated as black-box problems, i.e., the known global optima of these test functions cannot be used by the algorithms during the iterations, so as to compare the performance of these algorithms. The fitness value and the number of global optima of each function have been listed in Table 1. Here, we use fitness error $\varepsilon_f$, i.e., level of accuracy, to judge whether a solution is a real global optimum, i.e., if the difference between the fitness of a solution and the known global optimum is lower than $\varepsilon_f$, this solution can be considered as a global optimum. In our experiments, the fitness error $\varepsilon_f$, population size and maximal number of function evaluations for WSA and the 7 algorithms compared are listed in Table 2. It is worth to note that a function which has more optima or higher dimension requires a larger population size and more number of function evaluations.

Table 2. Test functions setting.

| Function no. | $\varepsilon_f$ | population size | No. of function evaluations |
|---|---|---|---|
| F1 | 0.01 | 100 | 10000 |
| F2 | 0.000001 | 100 | 10000 |
| F3 | 0.05 | 100 | 10000 |
| F4 | 0.001 | 100 | 10000 |
| F5 | 0.05 | 300 | 100000 |
| F6 | 0.002 | 200 | 20000 |
| F7 | 0.001 | 100 | 500000 |
| F8 | 0.001 | 100 | 500000 |
| F9 | 0.001 | 100 | 500000 |
| F10 | 0.001 | 100 | 500000 |
| F11 | 0.001 | 100 | 500000 |
| F12 | 0.001 | 100 | 500000 |

The user-specified control parameters of WSA, i.e., attenuation coefficient $\eta$, for these test functions are set as shown in Table 3.

Table 3. Parameter setting of WSA for test functions.

| Parameter | F1 | F2 | F3 | F4 | F5 | F6 | F7 | F8 | F9 | F10 | F11 | F12 |
|---|---|---|---|---|---|---|---|---|---|---|---|---|
| $\eta$ | 40 | 40 | 1.55 | 5.5 | 0.6 | 1.5 | 7.5E-3 | 2.2E-3 | 5E-3 | 6.5E-2 | 5E-3 | 6.5E-2 |

The parameters value of the 7 algorithms compared are set as the same as those in their reference source respectively. Table 4 has shown the setting of the main parameters of these algorithms. The parameter species radius $r_s$ of SDE and SPSO for these test functions are listed in Table 5.

Table 4. Setting of the parameters of algorithms.

| Algorithms | Parameters |
|---|---|
| GA | $P_c = 0.95$, $P_m = 0.05$ |
| DE | $P_c = 0.7$, $F = 0.5$ |
| PSO | $\omega = 0.729844$, $c_1 = 2$, $c_2 = 2$ |
| LIPS | $\omega = 0.729844$, $nsize = 2\sim5$ |
| SDE | $P_c = 0.9$, $F = 0.5$, $m = 10$ |
| CDE | $P_c = 0.9$, $F = 0.5$, $CF$ = population size |
| SPSO | $\chi = 0.729844$, $\varphi_1 = 2.05$, $\varphi_2 = 2.05$ |

1. $P_c$: crossover probability; $P_m$: mutation probability; 2. $F$: scaling factor; 3. $\omega$: inertia weight; $c_1$, $c_2$: acceleration factor; 4. $nsize$: neighborhood size; 5. $m$: species size; 6. $CF$: crowding factor; 7. $\chi$: constriction factor; $\varphi_1$, $\varphi_2$: coefficient.

Table 5. Species radius setting for test functions.

| | F1 | F2 | F3 | F4 | F5 | F6 | F7 | F8 | F9 | F10 | F11 | F12 |
|---|---|---|---|---|---|---|---|---|---|---|---|---|
| SDE | 0.05 | 0.05 | 1 | 0.5 | 1 | 1 | 4 | 10 | 4 | 1 | 4 | 1 |
| SPSO | 0.01 | 0.01 | 1 | 0.2 | 1.2 | 2 | 800 | 4000 | 800 | 120 | 800 | 60 |

## D. Performance metrics

To compare the performance of WSA with the 7 algorithms, we have conducted 25 independent runs for each algorithm on each test function. And the following five metrics are used to measure the performance of all the algorithms.

1) Success Rate (SR) [33]: the percentage of runs in which all the global optima are successfully located using the given level of accuracy.

2) Average Number of Optima Found (ANOF) [36]: the average number of global optima found over 25 runs.

3) Maximum Peak Ratio Statistic (MPR) [36]: this paper also adopts MPR to compare the quality of optima found by different algorithms. MRP is expressed as follows.

$$\text{MPR} = \frac{\sum_{i=1}^{q}(F_i - F^* + 1)}{\sum_{i=1}^{q}(f_i - F^* + 1)} \qquad (3)$$

where $q$ is the number of optima found by the algorithm, $\{f_i\}_{i=1}^{q}$ are the fitness value of these optima, $\{F_i\}_{i=1}^{q}$ are the value of real optima corresponding to those optima found by the algorithm, while $F^*$ is the value of the global optimum. It is obvious that the larger the MPR value is, the better the algorithm performs. The maximal MPR value is 1.

4) Convergence speed: the speed of an algorithm converging to the global optimum over function evaluations.

## IV. EXPERIMENTAL RESULTS AND ANALYSIS

This section presents and analyzes the results of comparative experiments. All the algorithms were run under the experiment setup shown in the previous section.

### A. Success rate

The success rates of all the algorithms on each test function are presented in Table 6, in which the numbers within parentheses denote the ranks of each algorithm. If the success rates of any two algorithms on a test function are equal, they have the same ranks over this test function. The last row of this table shows the total ranks of algorithms, which are the summation of the individual ranks on each test function. As we can see from Table 6, for multimodal functions F1-F10, the success rate of WSA on F3 is only a little bit lower than that of LIPS, but is far greater than those of other algorithms. Only two multimodal optimization algorithms (i.e., LIPS and SDE) can achieve nonzero success rates on F5, and no algorithm can achieve nonzero success rates on the four high dimensional multimodal functions F7-F10. What's more, WSA has achieved 100% success rates on test functions F1, F2, F4 and F6, which are much higher than those gained by most of other algorithms. Therefore, it can be seen that WSA has a very competitive performance on dealing with multimodal functions with respect to other algorithms. And for high-dimensional unimodal functions F11-F12, all the algorithms cannot achieve nonzero success rates on F12. However, WSA has achieved 100% success rate on F11, while the success rates of other algorithms on F11 are 0. Therefore, it can be concluded that WSA also has better performance than other algorithms on success rate when solving unimodal functions. It also can be seen that the better performance of WSA on success rate can be supported by the total rank of WSA that is 15 which is much smaller than those gained by other algorithms. The better performance of WSA is due to its novel iteration rules based on the behavior of whales hunting, including that the random movement of a whale is guided by its better and nearest whale, and its range of movement depends on the intensity of the ultrasound received as shown as Eq. 2, which have a great contribution to the maintenance of population diversity and the enhancement of global exploration ability, so as to locate the global optimum(optima).

Table 6. SR and ranks (in parentheses) of algorithms for test functions.

| Function no. | WSA | GA | DE | PSO | CDE | SDE | SPSO | LIPS |
|---|---|---|---|---|---|---|---|---|
| F1 | 1 | 0.96 | 1 | 1 | 1 | 1 | 0.32 | 0.64 |
|  | (1) | (6) | (1) | (1) | (1) | (1) | (8) | (7) |
| F2 | 1 | 0 | 0 | 0 | 0.04 | 0.24 | 0 | 0 |
|  | (1) | (4) | (4) | (4) | (3) | (2) | (4) | (4) |
| F3 | 0.8 | 0 | 0 | 0 | 0.08 | 0.5 | 0 | 1 |
|  | (2) | (5) | (5) | (5) | (4) | (3) | (5) | (1) |
| F4 | 1 | 0 | 0 | 0.08 | 0.28 | 0.52 | 0.12 | 1 |
|  | (1) | (7) | (7) | (6) | (4) | (3) | (5) | (1) |
| F5 | 0 | 0 | 0 | 0 | 0 | 0.16 | 0 | 0.76 |
|  | (3) | (3) | (3) | (3) | (3) | (2) | (3) | (1) |
| F6 | 1 | 0 | 0 | 0 | 0.04 | 0.24 | 0 | 0 |
|  | (1) | (4) | (4) | (4) | (3) | (2) | (4) | (4) |
| F7 | 0 | 0 | 0 | 0 | 0 | 0 | 0 | 0 |
|  | (1) | (1) | (1) | (1) | (1) | (1) | (1) | (1) |
| F8 | 0 | 0 | 0 | 0 | 0 | 0 | 0 | 0 |
|  | (1) | (1) | (1) | (1) | (1) | (1) | (1) | (1) |
| F9 | 0 | 0 | 0 | 0 | 0 | 0 | 0 | 0 |
|  | (1) | (1) | (1) | (1) | (1) | (1) | (1) | (1) |
| F10 | 0 | 0 | 0 | 0 | 0 | 0 | 0 | 0 |
|  | (1) | (1) | (1) | (1) | (1) | (1) | (1) | (1) |
| F11 | 1 | 0 | 0 | 0 | 0 | 0 | 0 | 0 |
|  | (1) | (2) | (2) | (2) | (2) | (2) | (2) | (2) |
| F12 | 0 | 0 | 0 | 0 | 0 | 0 | 0 | 0 |
|  | (1) | (1) | (1) | (1) | (1) | (1) | (1) | (1) |
| Total rank | 15 | 36 | 31 | 30 | 25 | 20 | 36 | 25 |

As some algorithms cannot obtain nonzero success rates on some multimodal functions, the metric ANOF has been used to test the performance of those algorithms on locating multiple global optima. Table 7 has presented the ANOF of all the algorithms over functions F2-F6 which have multiple global optima. As can be seen from this table, for these multimodal functions, the ANOF of WSA on test functions F2, F4 and F6 are much higher than those obtained by most of other algorithms, which echoes the 100% success rates of WSA on these functions as shown in Table 6. And the ANOF of WSA on F3 is only a little bit lower than that of LIPS, but is much higher than those of other algorithms, which is similar to the case of success rates of algorithms on this function. As can be seen from Table 6, only LIPS and SDE can achieve nonzero success rates on F5. Here, the ANOF of WSA on F5 is 6.76, which is much higher than those of other algorithms but the multimodal optimization algorithms LIPS and SDE. Therefore, the results of Table 7 have further demonstrated the outstanding performance of WSA on finding multiple global optima with respect to other algorithms when solving multimodal functions.

Table 7. ANOF and ranks (in parentheses) of algorithms for test functions F2-F6.

| Function no. | WSA | GA | DE | PSO | CDE | SDE | SPSO | LIPS |
|---|---|---|---|---|---|---|---|---|
| F2 | **5** | 1.04 | 1.08 | 1.64 | 2.72 | 3.32 | 1 | 1 |
|  | **(1)** | (6) | (5) | (4) | (3) | (2) | (7) | (7) |
| F3 | 3.8 | 0.08 | 1 | 1.16 | 2.72 | 2.92 | 1 | **4** |
|  | (2) | (8) | (6) | (5) | (4) | (3) | (6) | **(1)** |
| F4 | **2** | 0.36 | 1 | 1.08 | 0.64 | 1.52 | 1.04 | **2** |
|  | **(1)** | (8) | (6) | (4) | (7) | (3) | (5) | **(1)** |
| F5 | 6.76 | 0.16 | 2.16 | 1.44 | 1.44 | 11.84 | 1.04 | **17.72** |
|  | (3) | (8) | (4) | (5) | (5) | (2) | (7) | **(1)** |
| F6 | **3** | 0 | 1 | 1.16 | 1.08 | 1.92 | 0 | 1 |
|  | **(1)** | (7) | (5) | (3) | (4) | (2) | (7) | (5) |
| Total rank | 8 | 37 | 26 | 21 | 23 | 12 | 32 | 15 |

*B. Quality of optima found*

MPR is utilized to measure the quality of optima found by algorithms. The mean and standard deviation of MPR of all the algorithms on each test function over 25 runs are listed in Table 8. Here, the ranks of algorithms are based on the mean of MPR over the test functions. As we can see from this table, for low-dimensional multimodal functions F1-F6, WSA has achieved the best MPR on F1 and F2. And DE algorithm ranks the best on F3-F6. But DE algorithm has not got nonzero success rates on F3-F6 as shown in Table 6, and has gained worse ANOF than most of other algorithms as shown in Table 7, which mean that DE algorithm has a poor performance on locating multiple global optima though it can achieve a few of the multiple global optima with high accuracy, when solving low-dimensional multimodal functions. Whereas, WSA has achieved very good MPR over F3-F6, on the premise of keeping excellent SR and ANOF as shown in Table 6 and Table 7. And for high-dimensional multimodal functions F7 and F8, WSA only performs a little bit worse than LIPS and SDE, but outperforms other algorithms. What's more, for high-dimensional multimodal functions F9-F10 and high-dimensional unimodal functions F11-F12, WSA has achieved the best MPR values when compared with all the other algorithms. Particularly, WSA has gained the maximal MPR value (i.e., 1) on F11, and the mean of the optimal value found by WSA on F11 over 25 runs is **2.6E-9** in the experiments, which is far better than those obtained by all the other algorithms. Furthermore, the standard deviations of MPR of WSA on these test functions are quite small. Therefore, it can be concluded that WSA also performs better than most of other algorithms in terms of the quality of optima. The outstanding performance of WSA on the quality of optima is also due to its novel iteration rules, which contribute significantly to enhancing the local exploitation ability.

Table 8. MPR and ranks (in parentheses) of algorithms for test functions.

| Function no. | Measure | WSA | GA | DE | PSO | CDE | SDE | SPSO | LIPS |
|---|---|---|---|---|---|---|---|---|---|
| F1 | Mean<br>Std. | **0.99911**<br>0<br>**(1)** | 0.997047<br>8.45E-03<br>(5) | **0.99911**<br>0<br>**(1)** | **0.99911**<br>0<br>**(1)** | 0.996549<br>2.72E-03<br>(6) | 0.99799<br>2.19E-03<br>(4) | 0.983132<br>1.47E-02<br>(8) | 0.991751<br>7.57E-03<br>(7) |
| F2 | Mean<br>Std. | **1**<br>0<br>**(1)** | 1<br>9.80E-07<br>(1) | 1<br>0<br>(1) | 1<br>0<br>(1) | 1<br>5.77E-07<br>(1) | 1<br>0<br>(1) | 0.999839<br>2.54E-04<br>(8) | 0.999893<br>1.56E-04<br>(7) |
| F3 | Mean<br>Std. | 0.999494<br>1.99E-03<br>(3) | 0.673912<br>2.00E-01<br>(8) | **1**<br>0<br>**(1)** | 0.99778<br>5.78E-03<br>(5) | 0.977414<br>1.48E-02<br>(7) | 0.999729<br>1.14E-04<br>(2) | 0.995146<br>1.19E-02<br>(6) | 0.997927<br>2.93E-03<br>(4) |
| F4 | Mean<br>Std. | 0.999958<br>8.67E-05<br>(4) | 0.988828<br>2.19E-02<br>(8) | **1**<br>0<br>**(1)** | 0.99998<br>8.54E-05<br>(3) | 0.998787<br>9.81E-04<br>(7) | 0.999106<br>6.28E-05<br>(6) | 0.999501<br>3.5E-04<br>(5) | 0.999984<br>3.71E-05<br>(2) |
| F5 | Mean<br>Std. | 0.998858<br>1.79E-03<br>(2) | 0.549757<br>3.42E-01<br>(8) | **1**<br>2.77E-07<br>**(1)** | 0.995392<br>7.49E-03<br>(4) | 0.935463<br>7.33E-02<br>(7) | 0.985102<br>9.30E-03<br>(5) | 0.955778<br>3.75E-02<br>(6) | 0.996966<br>1.90E-03<br>(3) |
| F6 | Mean<br>Std. | 0.999967<br>1.36E-04<br>(2) | 0.900287<br>1.02E-01<br>(7) | **1**<br>0<br>**(1)** | 0.999905<br>2.40E-04<br>(3) | 0.98708<br>5.67E-02<br>(6) | 0.998346<br>1.78E-04<br>(5) | 0.869836<br>7.82E-02<br>(8) | 0.99873<br>1.43E-03<br>(4) |
| F7 | Mean<br>Std. | 1.77e-04<br>5.07e-05<br>(3) | 6.04e-06<br>2.37e-07<br>(8) | 4.35e-05<br>9.03e-06<br>(5) | 7.17e-06<br>8.45e-07<br>(7) | 1.65e-05<br>7.88e-07<br>(6) | 5.90e-04<br>1.85e-05<br>(2) | 1.64e-04<br>1.29e-05<br>(4) | **6.41e-04**<br>1.29e-04<br>**(1)** |
| F8 | Mean<br>Std. | 4.25e-05<br>2.03e-06<br>(3) | 2.74e-05<br>1.04e-06<br>(7) | 3.01e-05<br>6.84e-07<br>(6) | 4.23e-05<br>4.27e-06<br>(4) | 2.62e-05<br>4.52e-07<br>(8) | 4.30e-05<br>2.78e-06<br>(2) | 3.59e-05<br>1.63e-06<br>(5) | **5.08e-05**<br>2.49e-06<br>**(1)** |
| F9 | Mean<br>Std. | **0.5**<br>4.56e-13<br>**(1)** | 1.84e-02<br>7.36e-04<br>(8) | 1.12e-01<br>2.74e-02<br>(5) | 2.47e-02<br>3.72e-03<br>(7) | 7.18e-02<br>2.87e-03<br>(6) | 4.57e-01<br>1.40e-03<br>(3) | 2.99e-01<br>1.32e-02<br>(4) | 4.63e-01<br>1.39e-02<br>(2) |
| F10 | Mean<br>Std. | **1.01e-02**<br>8.13e-05<br>**(1)** | 1.18e-08<br>1.00e-09<br>(8) | 2.17e-07<br>7.72e-08<br>(6) | 5.48e-08<br>3.26e-08<br>(7) | 7.55e-07<br>1.37e-07<br>(5) | 1.01e-04<br>8.83e-06<br>(3) | 1.54e-05<br>3.58e-06<br>(4) | 1.06e-04<br>6.33e-05<br>(2) |
| F11 | Mean<br>Std. | **1**<br>6.68e-09<br>**(1)** | 5.62e-06<br>2.84e-07<br>(8) | 3.99e-05<br>1.12e-05<br>(4) | 7.21e-06<br>8.43e-07<br>(6) | 1.79e-05<br>9.53e-07<br>(5) | 1.37e-03<br>7.30e-05<br>(2) | 4.73e-05<br>5.86e-06<br>(3) | 5.93e-06<br>4.98e-07<br>(7) |
| F12 | Mean<br>Std. | **3.72e-03**<br>6.49e-04<br>**(1)** | 1.81e-03<br>5.22e-04<br>(2) | 8.00e-04<br>7.79e-05<br>(5) | 4.43e-04<br>4.95e-05<br>(8) | 5.67e-04<br>3.23e-05<br>(7) | 1.63e-03<br>1.44e-04<br>(4) | 7.68e-04<br>3.44e-04<br>(6) | 1.78e-03<br>1.96e-04<br>(3) |
| Total rank | | 23 | 78 | 37 | 56 | 71 | 39 | 67 | 43 |

## C. Efficiency

Based on the previous, it can be seen that WSA has a quite competitive performance when compared with other algorithms, in terms of the location of multiple global optima and the quality of optima. This subsection discusses the efficiency of all the algorithms, mainly focus on the convergence speed.

*1) Convergence speed*

To further demonstrate the superiority of WSA, it is compared with other algorithms on F3 in terms of convergence speed in this subsection. The convergence curves of all the algorithms on F3 are depicted in Fig. 5, in which the abscissa values denote function evaluations and the ordinate values represent the average fitness values of population over 25 runs. As can be seen from Fig. 5, WSA converges slower than DE, LIPS and SDE in the early iterations. However, in the mid and later iterations, WSA converges faster than LIPS and SDE, and it can achieve a better value than LIPS and SDE do. Although DE algorithm can converge to the global optimum (-200), it can only locate one of the four global optima in a single run, as shown in Table 7. Therefore, it can be concluded that WSA has better performance in terms of convergence speed than other algorithms on the premise of keeping good SR and ANOF. The excellent performance of WSA on convergence speed is also due to its novel iteration rules based on the behavior of whales hunting as shown as Eq. 2.

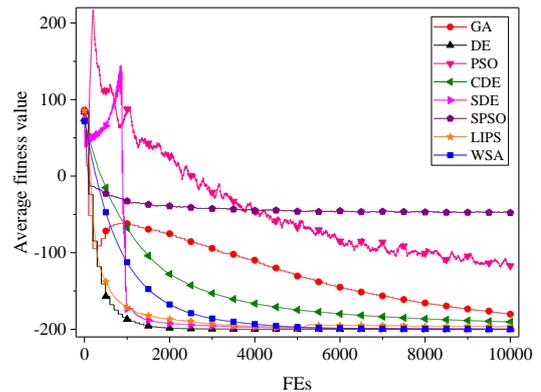

Fig. 5. The convergence graph of different algorithms on F3.

## V. Conclusions

A new swarm intelligence based metaheuristic called Whale Swarm Algorithm, inspired by the whales' behavior of communicating with each other via ultrasound for hunting, is proposed for function optimization in this paper. The innovations of the iterative equation of WSA consist of two parts: the random movement of a whale is guided by its better and nearest whale; and its range of movement depends on the intensity of the ultrasound received, which contribute significantly to the maintenance of population diversity, the avoidance of falling into the local optima quickly and the enhancement of global exploration ability, so as to locate the global optimum(optima). And the novel iteration rules also have a great contribution to the enhancement of local exploitation ability, especially when some whales have gathered around a same peak, so as to improve the quality of optima. WSA has been compared with several popular metaheuristic algorithms on four performance metrics (i.e., SR, ANOF, MPR and Convergence speed). The experimental results show that WSA has a quite competitive performance when compared with other algorithms, in terms of the location of multiple global optima, the quality of optima and efficiency.

In the future, we will focus on the following aspects:

1) Utilizing WSA to solve multi-objective optimization problems.

2) Modifying WSA to deal with real-world optimization problems, especially the discrete optimization problems and the NP-hard problems.

## Acknowledgment

This research was supported by the National Natural Science Foundation of China (NSFC) (51421062) and the National Key Technology Support Program (2015BAF01B04).

## References

[1] M. Mahi, Ö.K. Baykan, H. Kodaz, A new hybrid method based on Particle Swarm Optimization, Ant Colony Optimization and 3-Opt algorithms for Traveling Salesman Problem, Applied Soft Computing, 30 (2015) 484-490.
[2] K.C. Tan, Y. Chew, L.H. Lee, A hybrid multi-objective evolutionary algorithm for solving truck and trailer vehicle routing problems, European Journal of Operational Research, 172 (2006) 855-885.
[3] S.N. Qasem, S.M. Shamsuddin, S.Z.M. Hashim, M. Darus, E. Al-Shammari, Memetic multiobjective particle swarm optimization-based radial basis function network for classification problems, Information Sciences, 239 (2013) 165-190.
[4] B. Zeng, Y. Dong, An improved harmony search based energy-efficient routing algorithm for wireless sensor networks, Applied Soft Computing, 41 (2016) 135-147.
[5] E.S. Hou, N. Ansari, H. Ren, A genetic algorithm for multiprocessor scheduling, Parallel and Distributed Systems, IEEE Transactions on, 5 (1994) 113-120.
[6] B.-Y. Qu, P. Suganthan, S. Das, A distance-based locally informed particle swarm model for multimodal optimization, Evolutionary Computation, IEEE Transactions on, 17 (2013) 387-402.
[7] J. Holland, Adaptation in artificial and natural systems, Ann Arbor: The University of Michigan Press, (1975).
[8] M. Gen, R. Cheng, Genetic algorithms and engineering optimization, John Wiley & Sons, 2000.
[9] G. Syswerda, Schedule optimization using genetic algorithms, Handbook of genetic algorithms, (1991).
[10] T. Kellegöz, B. Toklu, J. Wilson, Comparing efficiencies of genetic crossover operators for one machine total weighted tardiness problem, Applied Mathematics and Computation, 199 (2008) 590-598.
[11] L.D. Whitley, Foundations of genetic algorithms 2, Morgan Kaufmann, 1993.
[12] Z. Michalewicz, Genetic algorithms+ data structures= evolution programs, Springer Science & Business Media, 2013.
[13] K. Deb, Multi-objective optimization using evolutionary algorithms, John Wiley & Sons, 2001.
[14] K. Deep, M. Thakur, A new mutation operator for real coded genetic algorithms, Applied mathematics and Computation, 193 (2007) 211-230.
[15] R. Storn, K. Price, Differential evolution-a simple and efficient adaptive scheme for global optimization over continuous spaces, ICSI Berkeley, 1995.
[16] A. Qing, Dynamic differential evolution strategy and applications in electromagnetic inverse scattering problems, Geoscience and Remote Sensing, IEEE Transactions on, 44 (2006) 116-125.
[17] Z. Gao, Z. Pan, J. Gao, A new highly efficient differential evolution scheme and its application to waveform inversion, Geoscience and Remote Sensing Letters, IEEE, 11 (2014) 1702-1706.
[18] S. Das, P.N. Suganthan, Differential evolution: a survey of the state-of-the-art, Evolutionary Computation, IEEE Transactions on, 15 (2011) 4-31.
[19] J. Kennedy, J.F. Kennedy, R.C. Eberhart, Y. Shi, Swarm intelligence, Morgan Kaufmann, 2001.
[20] M. Liu, S. Xu, S. Sun, An agent-assisted QoS-based routing algorithm for wireless sensor networks, Journal of Network and Computer Applications, 35 (2012) 29-36.
[21] R. Cheng, Y. Jin, A social learning particle swarm optimization algorithm for scalable optimization, Information Sciences, 291 (2015) 43-60.
[22] Y. Shi, R. Eberhart, A modified particle swarm optimizer, in: Evolutionary Computation Proceedings, 1998. IEEE World Congress on Computational Intelligence., The 1998 IEEE International Conference on, IEEE, 1998, pp. 69-73.
[23] Z.-H. Zhan, J. Zhang, Y. Li, H.S.-H. Chung, Adaptive particle swarm optimization, Systems, Man, and Cybernetics, Part B: Cybernetics, IEEE Transactions on, 39 (2009) 1362-1381.
[24] M. Dorigo, Optimization, learning and natural algorithms, Ph. D. Thesis, Politecnico di Milano, Italy, (1992).
[25] H. Drias, S. Sadeg, S. Yahi, Cooperative bees swarm for solving the maximum weighted satisfiability problem, in: Computational Intelligence and Bioinspired Systems, Springer, 2005, pp. 318-325.
[26] O.K. Erol, I. Eksin, A new optimization method: big bang–big crunch, Advances in Engineering Software, 37 (2006) 106-111.
[27] H. Zang, S. Zhang, K. Hapeshi, A review of nature-inspired algorithms, Journal of Bionic Engineering, 7 (2010) S232-S237.
[28] I. BoussaïD, J. Lepagnot, P. Siarry, A survey on optimization metaheuristics, Information Sciences, 237 (2013) 82-117.
[29] S. Majumdar, P.S. Kumar, A. Pandit, Effect of liquid-phase properties on ultrasound intensity and cavitational activity, Ultrasonics sonochemistry, 5 (1998) 113-118.
[30] V. Kenneth, Price, An introduction to differential evolution, New ideas in optimization, in, McGraw-Hill Ltd., UK, Maidenhead, UK, 1999.
[31] X. Li, Efficient differential evolution using speciation for multimodal function optimization, in: Proceedings of the 7th annual conference on Genetic and evolutionary computation, ACM, 2005, pp. 873-880.
[32] R. Thomsen, Multimodal optimization using crowding-based differential evolution, in: Evolutionary Computation, 2004. CEC2004. Congress on, IEEE, 2004, pp. 1382-1389.
[33] X. Li, Adaptively choosing neighbourhood bests using species in a particle swarm optimizer for multimodal function optimization, in: Genetic and Evolutionary Computation–GECCO 2004, Springer, 2004, pp. 105-116.
[34] K. Deb, Genetic algorithms in multimodal function optimization, Clearinghouse for Genetic Algorithms, Department of Engineering Mechanics, University of Alabama, 1989.
[35] J.-P. Li, M.E. Balazs, G.T. Parks, P.J. Clarkson, A species conserving genetic algorithm for multimodal function optimization, Evolutionary computation, 10 (2002) 207-234.
[36] http://www.ntu.edu.sg/home/EPNSugan/index_files/CEC2015/CEC2015.htm.